\pdfoutput=1
\documentclass[11pt]{article}

\usepackage[margin=1in]{geometry}
\usepackage{times}
\usepackage[T1]{fontenc}
\usepackage[utf8]{inputenc}
\usepackage{lmodern}        
\usepackage{microtype}

\usepackage{amsmath,amsfonts,bm}

\def\eqref#1{equation~\ref{#1}}

\def\1{\bm{1}}

\DeclareMathAlphabet{\mathsfit}{\encodingdefault}{\sfdefault}{m}{sl}
\SetMathAlphabet{\mathsfit}{bold}{\encodingdefault}{\sfdefault}{bx}{n}

\usepackage{hyperref}
\usepackage{url}
\usepackage{graphicx}
\usepackage{wrapfig}
\usepackage{hyperref}
\usepackage{url}
\usepackage{booktabs}
\usepackage{multirow}
\usepackage{graphicx}
\usepackage{xcolor}
\usepackage{amsmath}
\usepackage{xspace}
\usepackage{algorithm}
\usepackage{algpseudocode}
\usepackage{xspace}

\usepackage[numbers,sort&compress]{natbib} 
\usepackage{float}                         
\usepackage{authblk}                       

\def\methodname{Signed Quadratic Shrink\xspace}
\def\methodac{SQS\xspace}

\title{Preserving Bilinear Weight Spectra with a Signed and Shrunk Quadratic Activation Function}

\author[1]{Jason Abohwo}
\author[2]{Thomas Mosen}
\affil[1]{Yale University, USA}
\affil[2]{École polytechnique fédérale de Lausanne (EPFL), Switzerland}

\date{}

\begin{document}

\maketitle

\begin{abstract}
Understanding the inner workings of machine learning models is critical for ensuring their reliability and robustness. Whilst many techniques in mechanistic interpretability focus on activation-driven analyses, being able to derive meaningful features directly from the weights of a neural network would provide greater guarantees and more computational efficiency. Existing techniques for analyzing model features through weights suffer from drawbacks such as reduced performance and data inefficiency. In this paper, we introduce \methodname (\methodac)---an activation function designed to allow Gated Linear Units (GLUs) to learn interpretable features without these drawbacks. Our experimental results show that \methodac achieves performance competitive with state-of-the-art activation functions whilst enabling weight-based interpretability. 

\end{abstract}

\section{Introduction}
Understanding how neural networks make decisions is crucial for ensuring models are robust, and one promising avenue towards this is weight-based interpretability. One path to reaching weight-based interpretability that has been explored in previous work is to design performant, innately interpretable architectures. To this end, Sharkey \cite{sharkey2023technicalnotebilinearlayers} proposed the use of bilinear MLPs (MLPs in the form $(Wx + b) (Vx + c)$ i.e,. Gated Linear Units (GLUs) \cite{dauphin2017languagemodelinggatedconvolutional-glu-paper} without the activation function) to enable neural networks to learn interpretable features.

Following up on this, Pearce et al. \cite{pearce2024bilinearmlpsenableweightbased} demonstrated that Bilinear MLPs naturally lead to the development of features that can be interpreted via weight spectra. Though Bilinear MLPs allow for this, in some scenarios, they lag behind state-of-the-art GLUs like SwiGLU and GEGLU \cite{shazeer2020gluvariantsimprovetransformer} in performance \cite{shazeer2020gluvariantsimprovetransformer} and data efficiency\cite{shazeer2020gluvariantsimprovetransformer}. To rectify this shortcoming, we propose \methodname (\methodac)---an activation function for GLUs that yields performance and data efficiency on par with current activation functions (e.g., SwiGLU, GEGLU) \cite{touvron2023llama2openfoundation} whilst also preserving the weight structure properties that bilinear MLPs yield. 

\section{Related Works}

\subsection{Bilinear Multi-Layer Perceptrons}
Bilinear MLPs \cite{shazeer2020gluvariantsimprovetransformer, sharkey2023technicalnotebilinearlayers, pearce2024bilinearmlpsenableweightbased} are Gated Linear Units (GLUs)\cite{dauphin2017languagemodelinggatedconvolutional-glu-paper} that yield interpretable features due to a lack of activation function.
Traditional GLUs can be expressed as $(Wx + b) \odot \sigma(Vx + c)$, where $W$ and $V$ are weight matrices, $b$ and $c$ are vector biases, and $x$ is a given input vector. Bilinear MLPs are GLUs in the same form, without the activation function $\sigma$ (i.e. $(Wx + b) \odot (Vx + c)$).
From this, each output logit can be expressed as:
$$g(x)_a = (W_a^Tx)(V_a^Tx) \implies g(x)_a = (x^TW_a)(V_a^Tx) \implies g(x)_a = x^TA_ax$$

As stated by Pearce et al. \cite{pearce2024bilinearmlpsenableweightbased}, since the \textit{interaction matrix} $A_a$ is evaluated with two copies of x, it can be expressed as the sum of a symmetric and anti-symmetric matrix; however, the anti-symmetric component always evaluates to 0, which implies that $A_a$ is a symmetric matrix. As a result, through eigen-decomposition, we have the following:
$$g(x)_a = (x^TA_ax) = \sum_i^d \lambda_i(v_i^Tx)^2$$

\section{Methodology}

\subsection{\methodname}
To rectify the shortcomings of Bilinear MLPs, we introduce \methodname (\methodac). We motivate our activation function by the following properties of traditional MLPs and Bilinear MLPs. 

\begin{wrapfigure}{r}{0.5\linewidth} 
    \centering
    \includegraphics[width=\linewidth]{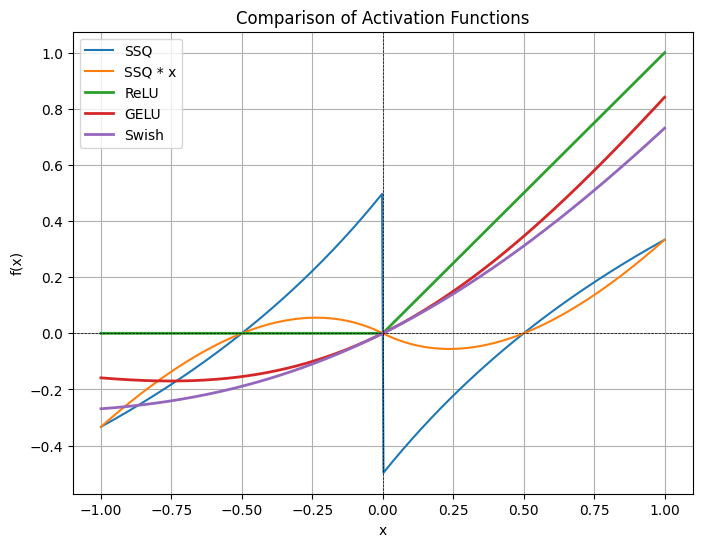}
    \caption{ReLU \cite{agarap2019deeplearningusingrectified-relu}, GeLU\cite{hendrycks2023gaussianerrorlinearunits-gelu}, Swish \cite{ramachandran2017searchingactivationfunctions-swish} \methodac($p = 1$, $\lambda = .5$, $c = .5$) functions. We include x $\cdot$ \methodac to provide intuition for the use of \methodac within a GLU\cite{dauphin2017languagemodelinggatedconvolutional-glu-paper}.}
    \label{fig:function_plotting}
\end{wrapfigure}

As noted by Pearce et al.\cite{pearce2024bilinearmlpsenableweightbased}, a given output logit of a bilinear layer can be formulated as $g(x)_a = (x^TA_ax) = \sum_i^d \lambda_i(v_i^Tx)^2$. With this, regular MLPs can be seen as a special case in which $\lambda$ and $v$ are shared between output logits. This then reduces the bilinear to an MLP with a quadratic activation function (i.e. $g(x) = ((XV)^{\odot 2})^T\Lambda$, where $V$ and $\Lambda$ are weight matrices). Quadratic activations have 2 main issues: vanishing gradients for small x and exploding gradients for large x. To rectify this, we modify the activation function as follows:

$$f(x) = x^2 \rightarrow g(x) = \frac{(|x| + c)^2 - c^2}{(1 + (\lambda |x|)^p)^{\frac{1}{p}}}$$

In this function, the $c$ hyperparameter provides a shift to the original function, and the term $(1 + (\lambda |x|)^p)^{\frac{1}{p}}$ adds a shrinking factor, governed by $\lambda$ and $p$, to the output. Note that $g = f$ if hyperparameters $c =\lambda = 0$ and $p = 1$. We then adapt this $g(x)$ for GLUs by factoring out an $|x|$ and adding a component to maintain directional information in the gate. Thus \methodname can be formulated as (we omit biases here for brevity):

$$
\sigma(x) =
\frac{x}{\lvert \lvert x \lvert \lvert}
\cdot
\frac{\lvert x \rvert - c}{(1 + (\lambda \lvert x\rvert)^p)^{\frac{1}{p}}} 
\quad \quad(Wx) \odot \sigma(Vx) = 
\frac{x}{\lvert \lvert x \lvert \lvert} \cdot
\frac{Wx \odot\lvert Vx \rvert - c(Wx)}{(1 + (\lambda \lvert Vx\rvert)^p)^{\frac{1}{p}}}
$$

Which can be approximated by the following if $p=1$:
$$
sgn(x) = \begin{cases}
1, & x \ge 0, \\
-1, & x < 0,
\end{cases}
\quad 
\sigma(x) = \frac{x - c\,\operatorname{sgn}(x)}{1 + \lambda\, x\,\operatorname{sgn}(x)}
$$
Effectively, this function is quasi-linear for $|x| << 1$ and $|x| > 10$ and is otherwise similar to a signed quadratic function. Figure \ref{fig:function_plotting} provides a visualization of \methodname.

\section{Experiments}
For our experiments, we evaluate \methodac, GeLU \cite{hendrycks2023gaussianerrorlinearunits-gelu}, and ReLU \cite{agarap2019deeplearningusingrectified-relu} GLUs as well as Bilinear GLUs \cite{pearce2024bilinearmlpsenableweightbased} on MNIST \cite{6296535-mnist}, Fashion MNIST \cite{xiao2017fashionmnistnovelimagedataset-fmnist}, and Tiny Stories \cite{eldan2023tinystoriessmalllanguagemodels}. For all uses of \methodac, we set $\lambda = .5$, $c=.01$ (as we find that this yields the most interpretable eigenvectors whilst maintaining performance on par with current activation functions), and $p = 1$ (for computational efficiency. We note that despite the complexity of the function, with $p = 1$, a triton implementation of \methodac runs in time on par---see Appendix C).

\subsection{MNIST and FMNIST: \methodac-GLU Learns Interpretable Eigenfeatures}
For this section, we follow the same procedure elucidated by Pearce et al. \cite{pearce2024bilinearmlpsenableweightbased}. More specifically, we analyze a shallow Feed-Forward network consisting of an embedding projection, an \methodac-GLU, and an output projection (see Appendix A for details). 
\begin{figure}
    \centering
    \includegraphics[width=.9\linewidth]{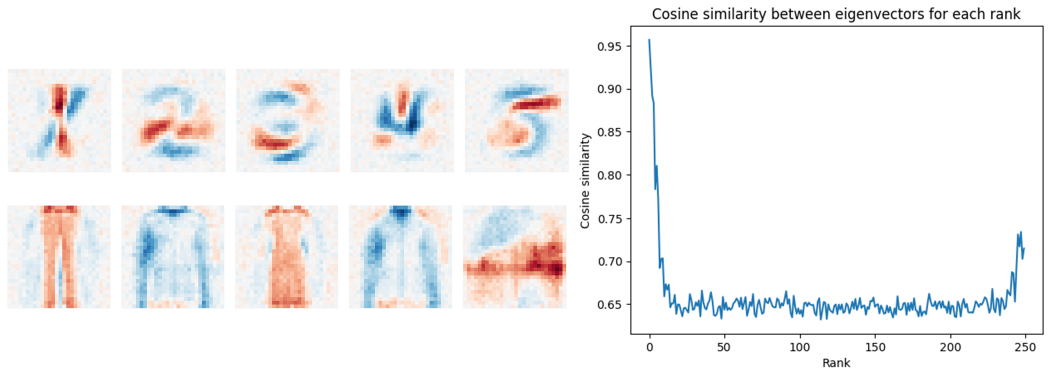}
    \caption{A) On the Top Eigenvectors after Eigendecomposition on each output dimension of the \methodac-GLU. The top row, in order, contains the eigenvectors for 1, 2, 3, 4, and 5. The bottom row, in order, contains the eigenvectors for trousers, pullover, dress, coat, and sandal. B) The cosine similarity between eigenvectors derived from an \methodac-GLU and a Bilinear MLP averaged across 5 runs.}
    \label{fig:eigenfeatures_cosinesim}
\end{figure}
Our results demonstrate that \methodac-GLU facilitates the learning of interpretable features as revealed through eigenvector decomposition. Figure \ref{fig:eigenfeatures_cosinesim} displays the top eigenvectors for five classes in both datasets. Visually, these eigenvectors can be seen to correspond to distinct classes in MNIST and FMNIST, respectively.

Additionally, we compare the eigenvectors derived from \methodac-GLU to those from Bilinear MLPs to assess similarity in learned features. As shown in Figure \ref{fig:eigenfeatures_cosinesim}, the cosine similarity between the eigenvectors from these two architectures is relatively high (never below .5 and .95 for the most important eigenvectors), supporting the notion that \methodac-GLU retains the interpretability of Bilinear MLPs. Through these experiments, we establish that \methodac-GLU, similarly to the Bilinear MLP \cite{pearce2024bilinearmlpsenableweightbased,} generates eigenfeatures that are meaningful and consistent with human intuition.

\subsection{Performance Results}
In this section, we demonstrate that \methodac achieves results competitive with previous activation functions with respect to final performance and data efficiency (measured as loss/evaluation metric vs training iteration). To this end, we conduct experiments on MNIST \cite{6296535-mnist}, FMNIST \cite{xiao2017fashionmnistnovelimagedataset-fmnist}, and Tiny-Stories \cite{eldan2023tinystoriessmalllanguagemodels}. 

\subsubsection{MNIST and FMNIST}
For MNIST \cite{6296535-mnist} and FMNIST \cite{xiao2017fashionmnistnovelimagedataset-fmnist}, we train 2-layer MLPs for 20 epochs with a batch size of 512 and report the average over 5 runs (more details in Appendix A). As shown by the graphs in Figure \ref{fig:per_iter}, \methodac both converges faster and achieves lower loss than both ReLU-GLUs and Bilinear MLPs, with a convergence rate near that of SwiGLU \cite{ramachandran2017searchingactivationfunctions-swish, shazeer2020gluvariantsimprovetransformer} and GELU \cite{hendrycks2023gaussianerrorlinearunits-gelu} whilst achieving loss lower than all other activation functions at the end of training. With respect to accuracy, the \methodac-GLU is the first to reach 80\%, 85\%, and 90\% accuracy, and achieves end-of-training accuracy on par with all other tested activation functions.

More quantitative details are summarized in Table \ref{performance_table}. For MNIST, while GELU and SwiGLU eventually attain slightly lower final losses (e.g., GELU reaching 0.0680 at the 100\% epoch compared to 0.0834 for \methodac), the \methodac-GLU outperforms both the ReLU-GLU and Bilinear MLP at all intermediate checkpoints. Similarly, on FMNIST, \methodac-GLU consistently shows improved loss reduction over ReLU and Bilinear MLPs, with its final loss being comparable to that of the best-performing activation functions. In addition, the final classification accuracy for \methodac-GLU is on par with these state-of-the-art activations on both MNIST and FMNIST.  

\subsubsection{Tiny Stories}
To evaluate the performance of \methodac-GLU on language modeling tasks, we conduct experiments on the Tiny Stories dataset \cite{eldan2023tinystoriessmalllanguagemodels}. For experimentation, we use a 4-layer Transformer architecture (further architectural and training details are provided in Appendix A). All models are trained with the same hyperparameter settings (e.g., learning rate, batch size, weight decay) and optimized using AdamW \cite{loshchilov2019decoupledweightdecayregularization-adamw} with a cosine annealing \cite{loshchilov2017sgdrstochasticgradientdescent-cosine} schedule over 3 epochs. As shown by the performance in \ref{performance_table} \methodac yields the best performance both in terms of loss and perplexity at the 50\%, 75\%, and 100\% steps. We display samples of model outputs from each model in Appendix A.

\begin{figure}
    \centering
    \includegraphics[width=1\linewidth]{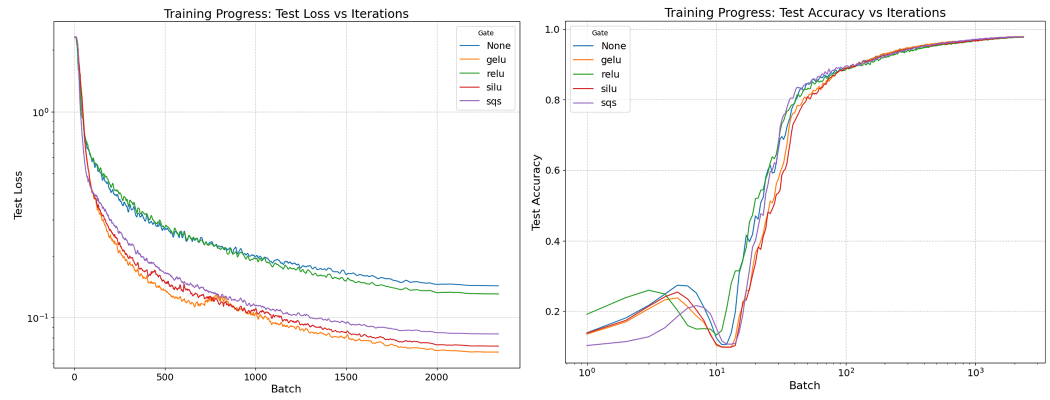}
    \caption{MNIST Test loss and accuracy over time. Note that log is applied to both graphs, to test loss on the left and to iteration on the right, and that "None" is the Bilinear MLP}
    \label{fig:per_iter}
\end{figure}

\begin{table*}[!htbp]
    \centering
    \label{performance_table}
    \caption{Performance of various activation functions after optimization steps. MNIST and FMNIST results are averaged over 5 runs. The best results for each category are bolded.}
    \resizebox{\textwidth}{!}{
    \begin{tabular}{c|c|cc|cc|cc|cc|}
    \toprule
    \multirow{2}{*} {\textbf{Dataset}} & 
    \multirow{2}{*} {\textbf{Activation}} & 
    \multicolumn{2}{c|}{\textbf{Step 25\% }} & 
    \multicolumn{2}{c|}{\textbf{Step 50\% }} & 
    \multicolumn{2}{c|}{\textbf{Step 75\% }} & 
    \multicolumn{2}{c|}{\textbf{Step 100\% }} \\ 
    
     &  &
     
     \textbf{Loss} & \textbf{Metric} & 
     \textbf{Loss} & \textbf{Metric} & 
     \textbf{Loss} & \textbf{Metric} & 
     \textbf{Loss} & \textbf{Metric} 
     \\

     \midrule
     
    \multirow{5}{*}
    {\textbf{MNIST}} 
    & ReLU & 0.2703 & 0.9550 & 0.1803 & 0.9690 & 0.1433 & 0.9748 & 0.1305 & 0.9770

    \\ 
    & GELU & \textbf{0.1305} & \textbf{0.9606} & \textbf{0.0943} & 0.9713 & \textbf{0.0747} & \textbf{0.9769} & \textbf{0.0680} & \textbf{0.9789}

    \\
    & SwiGLU & 0.1453 & 0.9573 & 0.1014 & 0.9695 & 0.0794 & 0.9757& 0.0728 & 0.9780

    \\
    & Bilinear & 0.2637 & 0.9578 & 0.1885 & 0.9703 & 0.1560 & 0.9755 & 0.1431 & 0.9780

    \\
    & \methodac & 0.1609 & 0.9590 & 0.1083 & \textbf{0.9722} & 0.0895 & 0.9768& 0.0834 & 0.9785

    \\ \midrule

    \multirow{5}{*}
    {\textbf{FMNIST}}
    & ReLU & 0.6852 & 0.7939 & 0.6099 & 0.8173 & 0.5546 & 0.8348 & 0.5322 & \textbf{0.8422}
    \\ 
    & GELU & \textbf{0.4723} & \textbf{0.8251} & 0.4109 & \textbf{0.8481} & \textbf{0.3746} & 0.8616 & \textbf{0.3599} & 0.8681
    \\
    & SwiGLU & 0.4731 & \textbf{0.8251} & \textbf{0.4102} & \textbf{0.8481} & 0.3747 & \textbf{0.8621} & 0.3605 & 0.8677
    \\
    & Bilinear & 0.7038 & 0.7961 & 0.6367 & 0.8211 & 0.5881 & 0.8383 & 0.5672 & 0.8461
    \\
    & \methodac & 0.5104 & 0.8217 & 0.4457 & 0.8422 & 0.4063 & 0.8548 & 0.3890 & 0.8616
    \\ \midrule

    \multirow{5}{*}
    {\textbf{Tiny Stories}}
    & ReLU & 2.1264 & 9.4803 & 1.9958 & 8.3943 & \textbf{1.9434} & 7.9114 & 1.9021 & 7.6538
    \\ 
    & GELU & 2.1257 & 9.4852 &  1.9980 & 8.5438 & 1.9471 & 8.0483 & 1.9061 & 7.8047
    \\
    & SwiGLU & 2.1227 & \textbf{9.3327} & 1.9945 & 8.2885 & 1.9456 & 7.9450 & 1.9042 & 7.7329
    \\
    & Bilinear & \textbf{2.1177} & 9.5730 & 1.9952 & 8.4898 & 1.9483 & 8.0583 & 1.9065 & 7.7987
    \\
    & \methodac & 2.1327 & 9.5220 & \textbf{1.9939} & \textbf{8.2721} & \textbf{1.9434} & \textbf{7.8579} & \textbf{1.9020} & \textbf{7.6286}
    \\ \midrule
    \end{tabular}
    }
\end{table*}

\section{Conclusion}
We have proposed \methodname---an activation function that allows for the development of interpretable features whilst providing performance on par, and in some scenarios, better, than standard activation functions.

\appendix

\section{Experimental Details}
This section details the experimental configurations used in our evaluation of \methodname (\methodac).

\subsection{Datasets}
\begin{itemize}
    \item MNIST: Dataset of 70,000 grayscale images (60,000 train, 10,000 test) of handwritten digits (0-9)
    \item Fashion MNIST: Dataset of 70,000 grayscale images (60,000 train, 10,000 test) of various wardrobe items (e.g. shirts, shoes, trousers)
    \item Tiny Stories: Dataset of ~2 Million (2.12 Million train, 22,000 test) short stories
\end{itemize}

\subsection{Hyperparameters}
\subsubsection{Section 4.1 Experimentation Hyperparameters}

\begin{table}[h!]
\centering
\caption{Training setup for the MNIST and FMNIST models in section 4.1.}
\begin{tabular}{ll}
\toprule
\textbf{MNIST Training Parameters} & \textbf{Values} \\
\midrule
layers           & 1 \\
input noise std  & 1.0 \\
weight decay     & 0.1 \\
learning rate    & 0.001 \\
batch size       & 2048 \\
optimizer        & AdamW \\
schedule         & cosine annealing \\
epochs           & 20 \\
\bottomrule
\end{tabular}
\end{table}
\subsubsection{Section 4.2 Experimentation Hyperparameters}

\begin{table}[h!]
\centering
\caption{Training setup for the MNIST and FMNIST models in section 4.2.}
\begin{tabular}{ll}
\toprule
\textbf{MNIST Training Parameters} & \textbf{Values} \\
\midrule
layers           & 2 \\
input noise std  & 1.0 \\
model dim        & 128 \\
weight decay     & 0.1 \\
learning rate    & 0.001 \\
batch size       & 512 \\
optimizer        & AdamW \\
schedule         & cosine annealing \\
epochs           & 20 \\
\bottomrule
\end{tabular}
\end{table}

\begin{table}[h!]
\centering
\caption{Training setup for the Tiny-Stories models.}
\begin{tabular}{ll}
\toprule
\textbf{Tiny-Stories Training Parameters} & \textbf{Values} \\
\midrule
layers           & 4 \\
model dim        & 128 \\
hidden dim       & 128 \\
weight decay     & 0.1 \\
learning rate    & 0.001 \\
batch size       & 512 \\
optimizer        & AdamW \\
schedule         & cosine annealing \\
epochs           & 20 \\
\bottomrule
\end{tabular}
\end{table}

\section{Additional Interpretability Results}
In this section, we present further analyses of the eigen-decomposition of the \methodac-GLU weights. Our goal is to deepen the understanding of how \methodac-GLU facilitates weight-based interpretability compared to conventional bilinear MLPs. 
\subsubsection{Eigen-Spectrum Analysis}
\begin{figure}
    \centering
    \includegraphics[width=1\linewidth]{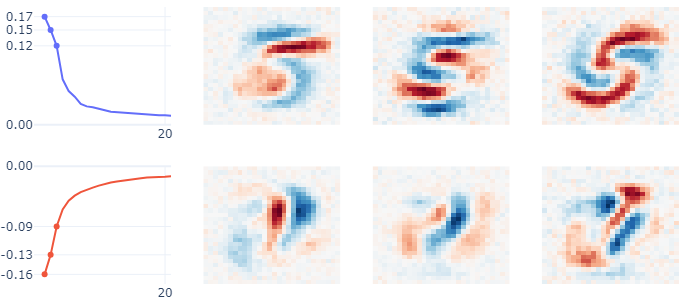}
    \caption{Eigen-spectrum for the 5 logit of the \methodname-GLU trained on MNIST}
    \label{fig:eigenspectrum5}
\end{figure}

\begin{figure}
    \centering
    \includegraphics[width=1\linewidth]{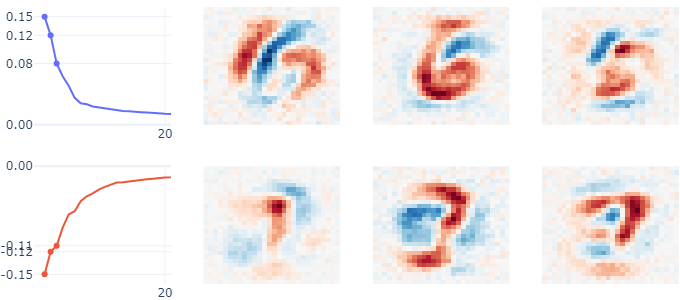}
    \caption{Eigen-spectrum for the 6 logit of the \methodname-GLU trained on MNIST}
    \label{fig:eigenspectrum6}
\end{figure}

\begin{figure}
    \centering
    \includegraphics[width=1\linewidth]{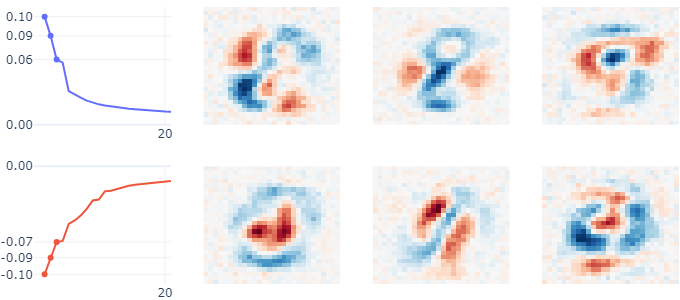}
    \caption{Eigen-spectrum for the 8 logit of the \methodname-GLU trained on MNIST}
    \label{fig:eigenspectrum8}
\end{figure}

\begin{figure}
    \centering
    \includegraphics[width=1\linewidth]{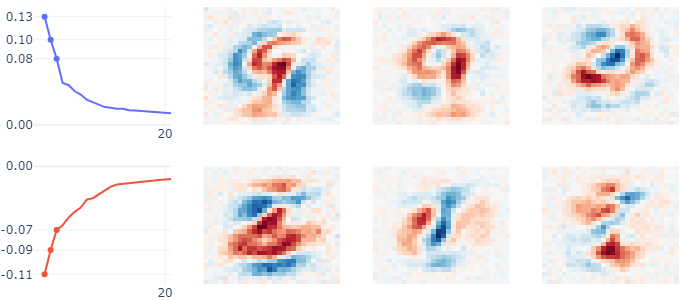}
    \caption{Eigen-spectrum for the 9 logit of the \methodname-GLU trained on MNIST}
    \label{fig:eigenspectrum9}
\end{figure}

Following the methodology of Pearce et al. \cite{pearce2024bilinearmlpsenableweightbased}, we compute the eigen-spectrum of the interaction matrices extracted from the \methodac-GLU layer. Figures \ref{fig:eigenspectrum5} \ref{fig:eigenspectrum6} \ref{fig:eigenspectrum8} \ref{fig:eigenspectrum8} \ref{fig:eigenspectrum9} show the eigenvalue distributions for several output classes on MNIST. Similarly to the properties that bilinear MLPs exhibit \cite{pearce2024bilinearmlpsenableweightbased}, the first few eigenvectors for a given output class are consistently significantly more important than other eigenvectors.
\begin{figure}
    \centering
    \includegraphics[width=1\linewidth]{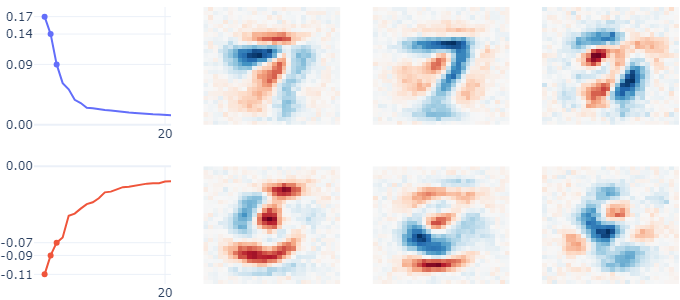}
    \caption{Eigen-spectrum for the 7 logit of the \methodname-GLU trained on MNIST}
    \label{fig:eigenspectrum7}
\end{figure}

\subsubsection{Eigenvectors as Explanations}
\begin{figure}
    \centering
    \includegraphics[width=1\linewidth]{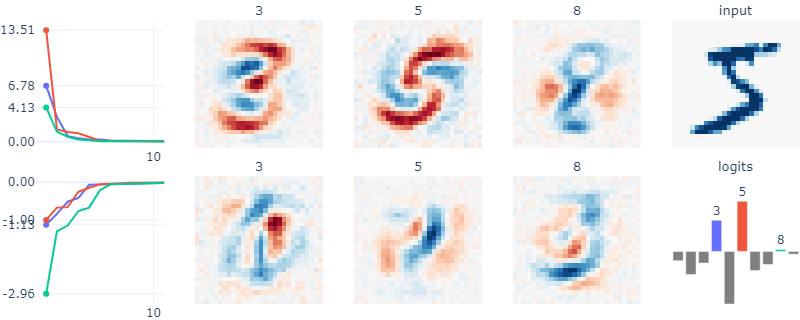}
    \caption{Caption}
    \label{fig:eigen_explanation}

\end{figure}

Similarly to Bilinear MLPs \cite{pearce2024bilinearmlpsenableweightbased}, we find that the eigenvectors of \methodac-GLUs can be used to explain why certain inputs were classified in certain ways, as demonstrated in figure \ref{fig:eigen_explanation}

\section{Computing \methodac}
In this section, we provide more details on how we compute \methodac and its speed relative to other activation functions. Algorithm \ref{alg:forward_pass} outlines how we compute the forward pass for \methodac, and Table \ref{tab:backward_time_horizontal} displays the forward + backward pass time for \methodac in comparison to other activation functions using a 1-layer GLU.

\begin{algorithm}[H]
\caption{\methodac Forward Pass}
\label{alg:forward_pass}
\begin{algorithmic}[1]
\Require Input tensor \(x\); parameters \(c \in \mathbb{R}\) (default \(0.01\)), \(\lambda \in \mathbb{R}\) (default \(0.5\)), and \(p\) (default \(1\))
\State \textbf{// Note: parameter \(p\) is defined but not used in the forward pass.}
\ForAll{elements \(x_i\) in \(x\)}
    \State \( s_i \gets \begin{cases}
            1.0, & \text{if } x_i \ge 0 \\
            -1.0, & \text{if } x_i < 0
         \end{cases} \)
\EndFor
\State Compute the output tensor element-wise:
\[
y = \frac{x - c \cdot s}{\,1 + \lambda \cdot x \cdot s\,}
\]
\Return \(y\)
\end{algorithmic}
\end{algorithm}

\begin{table}[htbp]
\centering
\begin{tabular}{r r r | r r r r r}
\toprule
Batch & \(d_{in}\) & \(d_{out}\) & ReLU & SwiGLU & GELU & Identity & \methodac \\
\midrule
256  & 256  & 256  & 0.5516 & 0.5617 & 0.6310 & 0.4941 & 0.5753 \\
256  & 512  & 512  & 0.5993 & 0.5765 & 0.5934 & 0.4984 & 0.5755 \\
256  & 1024 & 1024 & 0.6804 & 0.6914 & 0.6903 & 0.6663 & 0.6417 \\
256  & 2048 & 2048 & 2.5219 & 2.6016 & 2.6284 & 2.5625 & 2.3355 \\
512  & 256  & 256  & 0.5889 & 0.6403 & 0.5591 & 0.5053 & 0.5688 \\
512  & 512  & 512  & 0.5535 & 0.5563 & 0.5300 & 0.4947 & 0.5822 \\
512  & 1024 & 1024 & 1.2429 & 1.2648 & 1.2655 & 1.2195 & 1.1716 \\
512  & 2048 & 2048 & 4.5534 & 4.6016 & 4.6078 & 4.5101 & 4.3259 \\
1024 & 256  & 256  & 0.5610 & 0.5525 & 0.5537 & 0.5097 & 0.5827 \\
1024 & 512  & 512  & 0.7347 & 0.7535 & 0.7549 & 0.7105 & 0.6834 \\
1024 & 1024 & 1024 & 2.7023 & 2.7947 & 2.8036 & 2.6920 & 2.5393 \\
1024 & 2048 & 2048 & 8.4728 & 8.5362 & 8.5190 & 8.3451 & 8.0269 \\
2048 & 256  & 256  & 0.5717 & 0.5603 & 0.5627 & 0.5106 & 0.5894 \\
2048 & 512  & 512  & 1.4109 & 1.4287 & 1.4314 & 1.3428 & 1.3002 \\
2048 & 1024 & 1024 & 5.0301 & 5.1295 & 5.1456 & 5.0276 & 4.7280 \\
2048 & 2048 & 2048 & 16.8498 & 16.9756 & 16.9970 & 16.6761 & 16.0520 \\
\bottomrule
\end{tabular}
\caption{Average forward + backward time (ms) for various activations and configurations. Note, we use $\lambda = .5$, $c=.01$, $p=1$}
\label{tab:backward_time_horizontal}
\end{table}

\section{Tiny Stories Model Outputs}
In this section, we present example tiny stories outputs from each model (note: we manually clean unintelligible symbols from text that has them)

Gate: relu - Generated text:
\begin{quote}
once upon a time's car was in a sunny town wearing a unique red part. outside, there was a gloomy day. the rain stopped, the lights on the tracks and all the smoke were inside. the car was sad because it did not fit. the yelped and cried and cry all day. luckily, a kind lady noticed the gas conductor on the driveway. she said, that's normal. i can provide you with an expensive toy to supply again. the singer gave the danny a different type of bath. 
\end{quote}

Gate: SwiGLU - Generated text:
\begin{quote}
once upon a time a big room full of toys. the room was big and sparkly and had kept clay on top. a little girl named lily went to the kitchen to play. she saw a big table near the table. she thought it was yummy! lots of kids splashed and eat. lily's mommy saw her getting ready. she took the chairs with her and they all got some polish. lily was frustrated. she wanted to help. she took some soap and put the soap around it. it made her feel like she liked her tea. she played for hours and held a cloth together. 

\end{quote}

Gate: gelu - Generated text:
\begin{quote}
once upon a time up a room that was filled with stars a beautiful, dark night. there were potatoes and bugs were a happy little girl. she loved them there every night and couldn't wait to go outside and play. one day the famous seal came across a wide open field. along the way up to a solid mountain! the little girl was scared to stop fighting. but something was wrong that was wrong. she missed her wandering around. she decided that she found a soft spot under her bed as fast as jenny. the deep, the little girl could not help but welcome to the forest. 
\end{quote}

Gate: \methodac - Generated text:
\begin{quote}
once upon a time summer. her name was sarah. she was very brave. she wanted to stay healthy. one day, sarah decided to take a visit in the big city. she looked for works finally she saw the place. she slowly put her old clothes in the city for a long time. she was legs and loved to reach ships. sarah was so proud of herself for helping. she held the clothes in her hands and started to oinkly they talked about to step. her fingers parted the clothes and her beloved welcome, feeling a more worries worm so that his eyes wrote him on her skin. 
\end{quote}

Gate: None (i.e. Bilinear) - Generated text:
\begin{quote}
once upon a time a time, there was a little girl named lily. she loved to play outside and run around in the grass. one day, while playing in the park, lily saw a funny man. he was wearing a regular sandy suit and didn't know any person. lily asked her mom, " what is that man doing here? " her mom smiled and told her that she could go to the park, but she didn't want to play with him. she asked if she wanted to play, but her mom said there was only a princess who was ignorant.
\end{quote}

\end{document}